\documentclass[11pt]{article}

\usepackage[preprint]{acl}

\usepackage{times}
\usepackage{latexsym}

\usepackage[T1]{fontenc}

\usepackage[utf8]{inputenc}

\usepackage{microtype}

\usepackage{inconsolata}

\usepackage{graphicx}

\usepackage{booktabs}

\usepackage{graphicx}
\graphicspath{{figure-paper/}}

\usepackage{amsmath}
\usepackage{amssymb}
\usepackage{booktabs}
\usepackage{array}
\usepackage{tabularx}

\usepackage[table]{xcolor}
\definecolor{headerblue}{HTML}{DCE9F5}

\newcommand{\boldheader}[1]{\par\noindent\textbf{#1.}}
\title{When Do Language Models Add Value? Relative Competence in Hybrid Event Forecasting}

\author{
  \textbf{Aditi Tiwari} \and
  \textbf{Aashrith Bandaru} \and
  \textbf{Heng Ji} \\
  University of Illinois Urbana-Champaign \\
  Urbana, Illinois, USA \\
  \texttt{aditit5@illinois.edu, hengji@illinois.edu}
}

\begin{document}
\maketitle

\begin{abstract}
When a forecasting system already has a prediction from a market, crowd, or statistical model, when should it use a language model? Standalone accuracy cannot answer this question because a less accurate model may still improve the existing forecast. We study this distinction through relative competence, separating a model's replacement advantage from its complementary value. Under Brier loss (squared error between predicted probabilities and binary outcomes), we characterize when model disagreement can correct external forecast errors and derive the population gain from regime-specific affine pooling. We introduce a competence gate that estimates source weights from resolved outcomes, shrinks regime-specific estimates toward a global solution, and chronologically recalibrates the pooled forecast. Across 4,974 resolved binary questions and four language models, the best standalone model has 24.94\% higher Brier loss than the calibrated external forecast, yet the gate improves that reference by 3.35\% (95\% CI, 1.80--4.90\%). Model contributions vary across forecasting regimes, with larger corrections in thinner markets and forecasting crowds and smaller corrections in higher-participation settings. The gate also shows no statistically detectable performance difference from more flexible contextual aggregators. These results demonstrate that standalone superiority is not necessary for complementary predictive value and that language models should be evaluated relative to the forecasts already available.
\end{abstract}

\section{Introduction}
\label{sec:introduction}

\begin{figure*}[t]
\centering
\includegraphics[width=\textwidth]{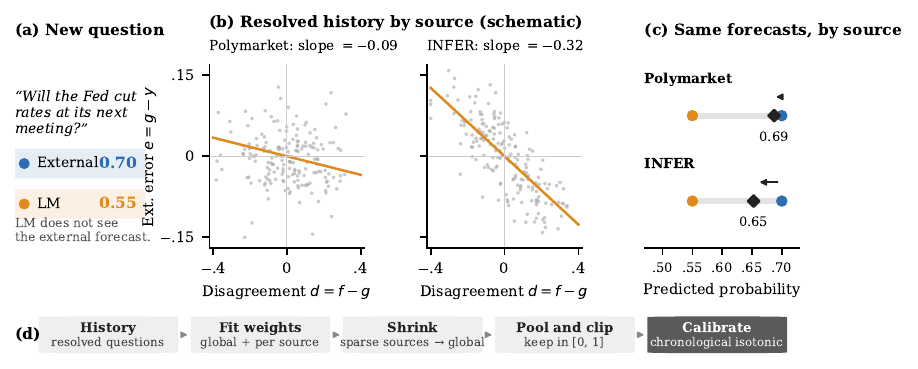}
\caption{Overview. (a) A question receives an external forecast and an independently elicited LM forecast. (b) Schematic resolved histories illustrate how the relationship between external error and model disagreement can differ across source-defined regimes. (c) The same hypothetical forecasts illustrate different source-specific corrections before calibration. (d) The gate fits global and source-specific weights from resolved outcomes, shrinks sparse estimates toward the global solution, then pools, clips, and chronologically calibrates forecasts.}
\label{fig:competence_gate}
\end{figure*}

Consider forecasting whether the Federal Reserve will cut interest rates at its next meeting. Suppose a prediction market assigns a 70\% probability, while a language model (LM) predicts 55\%. Should a forecasting system use the LM's prediction? The answer depends not only on which source is more accurate, but also on whether their errors differ in useful ways. A less accurate LM may still identify information missing from the market forecast. This question matters because forecasting systems often already have predictions from markets, human crowds, or statistical models \cite{wolfers2004prediction,mellers2014psychological,satopaa2014timeseries, halawi2024approaching}. The deployment problem is therefore not simply whether an LM forecasts well, but whether it improves the prediction already available.

Forecast combination has long studied how differently informed predictions can reduce error \cite{bates1969combination,granger1984improved,degroot1983comparison,strahl2017cross}. Learned aggregation methods, including stacking, extend this framework by optimizing how predictive distributions are combined \cite{yao2018stacking}. Recent hybrid systems also combine model-generated forecasts with human or market predictions and report improvements over individual sources \cite{benjamin2023hybrid,alur2025aia,kim2026forecasting}. However, aggregate improvements do not establish when an LM should influence an existing forecast or when that forecast should be largely preserved. The distinction matters because a market, forecasting crowd, and statistical model may already capture different amounts of the information available to an LM. We therefore study how the incremental value of LM predictions varies across forecasting regimes and whether that variation can be estimated from resolved outcomes.

We formalize this question through \emph{relative competence}, the predictive value of an LM conditional on an available external forecast. We distinguish \emph{replacement advantage}, which measures whether the LM is more accurate than the external source, from \emph{combination value}, which measures whether their forecasts can be combined to reduce error. Under Brier loss (squared error between predicted probabilities and binary outcomes), we show that complementary value depends on whether LM disagreement systematically corrects external forecast errors \cite{brier1950verification,gneiting2007strictly}. We also derive the population gain from regime-specific rather than global affine pooling. This characterization explains why standalone superiority is not necessary for complementary value and why different external sources can favor different model contributions.

We introduce a \emph{competence-gated forecast pool} that estimates source coefficients from previously resolved questions. The method fits global and regime-specific affine combinations, shrinks estimates from sparse regimes toward the global solution, and recalibrates the pooled forecast using chronological out-of-sample predictions \cite{zadrozny2002transforming}. The LMs forecast independently of the external probability, allowing their additional information to be evaluated without first exposing them to that source. Figure~\ref{fig:competence_gate} illustrates the method. Unlike self-reported confidence, the fitted coefficients reflect historical forecasting errors and provide an inspectable account of how each source contributes to the combination.

We evaluate four LMs from the gpt-oss \cite{gptoss} and Llama \cite{ grattafiori2024llama3} families on 4,974 resolved binary questions spanning markets, forecasting crowds, and structured forecasting tasks. The best standalone LM has 24.94\% higher Brier loss than the calibrated external forecast, yet the competence gate achieves a 3.35\% improvement over the external reference. Matched regime-specific pooling also improves on global affine pooling, although this smaller routing gain is secondary. More flexible contextual aggregators achieve similar predictive accuracy. Importantly, model-side corrections decrease as market participation or crowd size increases. One possible explanation is that broader external aggregation leaves less complementary information for the LM, although participation does not directly measure forecast accuracy. Independent elicitation also preserves complementary information: revealing the external forecast reduces measured combination value from 0.00153 to 0.00072. Historical comparative competence also improves on verbal confidence in a separate diagnostic \cite{xiong2024can}. Together, these findings show that useful model influence depends on the competing information source, rather than standalone accuracy or confidence alone.

Our contributions are threefold.
\vspace{-0.3em}
\begin{enumerate}
    \item We provide controlled evidence that individually weaker LMs improve stronger external forecasts, and that this complementary value varies across sources and participation levels and roughly halves when the LM sees the external forecast.
    \vspace{-0.2em}
    \item We formalize relative competence, separating replacement advantage from combination value, and characterize when disagreement corrects external forecast errors under Brier loss.
    \vspace{-0.2em}
    \item We develop a chronological competence-gated pool that estimates interpretable, shrinkage-stabilized source corrections from resolved outcomes.
\end{enumerate}

\section{Related Work}

\boldheader{LLM forecasting and hybrid systems}
Retrieval, aggregation, and post-cutoff evaluation have improved LLM forecasting~\citep{halawi2024approaching,schoenegger2024wisdom,turtel2025llms,karger2025forecastbench,cheng2026polybench}. Hybrid systems combine LM forecasts with markets or crowds~\citep{alur2025aia,kim2026forecasting,barot2026polyswarm}. \citet{alur2024human} study complementary human expertise in algorithmic prediction. We instead ask when an LM adds information beyond an existing external forecast. Our question-only, retrieval-free forecasts isolate this contribution from search quality.

\boldheader{Forecast combination and conditional pooling}
Classical work studies forecast combination and encompassing~\citep{bates1969combination,clemen1989combining,ranjan2010combining,chong1986encompassing,harvey1998encompassing}. Conditional predictive ability examines performance given an information set~\citep{giacomini2006tests}. Weak standalone models can also contribute through aggregation~\citep{peng2026twoheads}. Contextual and hierarchical pooling allow weights to vary~\citep{yao2022bayesian,oelrich2024local,johnson2018bayesian,mcalinn2020multivariate}. Our contribution is not a new pooling algorithm. We characterize complementary value under Brier loss and quantify the gain from regime-specific rather than global affine pooling.

\boldheader{Confidence, selective prediction, and deferral}
Confidence methods assess model uncertainty~\citep{xiong2024can}. Selective prediction and deferral study withholding or handing off predictions~\citep{mozannar2020consistent,geifman2017selective,cortes2016learning}. Forecast evaluation also requires leakage control~\citep{paleka2025pitfalls}. We instead evaluate comparative competence: whether historical signals identify when model information improves an external forecast, both in combination and selective ranking.

\section{Problem Setup}
\label{sec:problem}

\boldheader{Forecasting Setting}
We study binary event forecasting when a system already has a probability estimate from an external source. Question $i$ has text $x_i$, forecast time $t_i$, resolved outcome $y_i\in\{0,1\}$, and outcome-availability time $s_i$. The system has an external forecast $g_i\in[0,1]$ and forecasts $f_{im}\in[0,1]$ from $M$ language models. External forecasts may come from prediction markets, forecasting crowds, or statistical models. In the primary protocol, a language model sees the question, forecast date, and resolution criteria but not $g_i$.

Each question has an observed regime $k_i\in\mathcal K$. We use external-source identity as the primary regime because it is known when the forecast is made. Define the $S=M+1$ source probabilities as
\vspace{-0.5em}
\begin{equation}
\label{eq:source_vector}
\mathbf p_i=[g_i,f_{i1},\ldots,f_{iM}]^\top.
\end{equation}

We evaluate probabilities using Brier loss \cite{brier1950verification,gneiting2007strictly}, with population and regime-specific risks
\vspace{-0.5em}
\begin{equation}
\label{eq:brier_risk}
\begin{aligned}
\mathcal B(h)&=\mathbb E[(h-y)^2],\\
\mathcal B_k(h)&=\mathbb E[(h-y)^2\mid k].
\end{aligned}
\end{equation}

Lower risk is better. The calibrated external comparator is denoted $g_i^{\mathrm{cal}}$. We use raw $g_i$ to analyze disagreement and replacement advantage, but compare the deployed gate against chronological $g_i^{\mathrm{cal}}$ so improvements due only to external calibration are not attributed to LMs.

\boldheader{Relative Competence}
We use \emph{relative competence} to describe the predictive value of a language model given an external forecast. Two targets answer different questions.

\textbf{Replacement advantage} measures whether a model should be used instead of the external probability:
\vspace{-0.5em}
\begin{equation}
\label{eq:replacement_advantage}
\Delta_k^{\mathrm{rep}}=\mathcal B_k(g)-\mathcal B_k(f).
\end{equation}

A positive value indicates lower standalone model risk. Replacement advantage does not measure whether the model can improve a combination.

\textbf{Combination value} measures the best population improvement from an affine adjustment of $g$ using disagreement with $f$. Following linear forecast combination \cite{bates1969combination,granger1984improved}, define
\vspace{-0.5em}
\begin{equation}
\label{eq:two_source_pool}
h_\lambda=g+\lambda(f-g),\qquad \lambda\in\mathbb R.
\end{equation}
\vspace{-0.2em}
Its regime-specific combination value is
\vspace{-0.2em}
\begin{equation}
\label{eq:combination_value}
\Delta_k^{\mathrm{comb}}=\mathcal B_k(g)
-\inf_{\lambda\in\mathbb R}\mathcal B_k(h_\lambda).
\end{equation}

The unconstrained affine score is an analytic device and need not lie in $[0,1]$. The deployed estimator clips before returning a probability. Combination value can be positive even when replacement advantage is negative. Thus, the deployment question is not simply which forecaster has lower standalone error, but whether model disagreement helps correct errors in the available forecast.

With multiple models, an affine pool has score $\mathbf p_i^\top\mathbf w$ subject to $\mathbf1^\top\mathbf w=1$. Individual coefficients may be signed. The theory first studies one model to make the source-relative gain explicit, then the method fits all $S$ sources.

\section{Relative Competence Under Brier Loss}
\label{sec:theory}

\boldheader{When Disagreement Adds Value}
Let $d=f-g$ denote disagreement and $e=g-y$ denote external error. In regime $k$, write $A_k=\mathbb E_k[d^2]$ and $M_k=\mathbb E_k[de]$. Expanding the affine Brier risk gives
\vspace{-0.3em}
\begin{equation}
\footnotesize
\label{eq:affine_risk}
\mathcal R_k(\lambda)=\mathbb E_k[(e+\lambda d)^2]=\mathcal B_k(g)+2\lambda M_k+\lambda^2A_k.
\end{equation}

For $A_k>0$, the optimal coefficient and improvement are
\vspace{-0.3em}
\begin{equation}
\footnotesize
\label{eq:optimal_affine}
\lambda_k^\star=-\frac{M_k}{A_k},\qquad \Delta_k^{\mathrm{comb}}=\frac{M_k^2}{A_k}.
\end{equation}
When $A_k=0$, the two forecasts agree almost surely in regime $k$. We set $\lambda_k^\star=0$ by convention, and the regime contributes no combination value or routing gain.

When $M_k<0$, a positive model coefficient moves the pool in the direction of the model forecast. When $M_k>0$, a negative coefficient moves it in the opposite direction. If $M_k=0$, the optimal affine adjustment is zero. Disagreement magnitude alone therefore does not establish complementary value. The relevant quantity is its signed relationship with external forecast errors.

\boldheader{Gain from Regime-Specific Pooling}
Let $\pi_k=\Pr(k)$ and suppose $\sum_k\pi_kA_k>0$. The optimal single global coefficient is
\vspace{-0.3em}
\begin{equation}
\footnotesize
\label{eq:optimal_global_weight}
\lambda_G^\star=\frac{\sum_k\pi_kA_k\lambda_k^\star}{\sum_k\pi_kA_k}.
\end{equation}
\textbf{Proposition 1 (Affine routing identity).} The reduction in population risk obtained by using regime-specific optimal coefficients instead of the optimal global coefficient is
\vspace{-0.3em}
\begin{equation}
\footnotesize
\label{eq:routing_identity}
\sum_k\pi_k\mathcal R_k(\lambda_G^\star)-\sum_k\pi_k\mathcal R_k(\lambda_k^\star)=\sum_k\pi_kA_k(\lambda_k^\star-\lambda_G^\star)^2.
\end{equation}
The identity follows by completing the square around each regime optimum. It shows that regime-specific pooling can outperform one global coefficient when the regimes prefer different corrections. It does not guarantee that estimated regime weights improve out-of-sample forecasts. Proposition~1 concerns the unregularized population affine score before clipping and recalibration. We give the proof and multi-source extension in Appendix~\ref{app:theory}.

\section{Competence-Gated Forecast Pool}
\label{sec:method}
We learn a global source combination and regime-specific corrections from resolved outcomes, then shrink each regime estimate toward the global estimate. This keeps the estimator interpretable while limiting variance in small regimes. Figure~\ref{fig:competence_gate} illustrates the architecture.

\boldheader{Chronological Weight Estimation}
At forecast time $t_i$, only earlier questions with available outcomes and admissible source forecasts can be used:
\vspace{-0.3em}
\begin{equation}
\footnotesize
\label{eq:historical_set}
\mathcal H_i=\{j:s_j<t_i,\ \mathbf p_j\text{ admissible}\}.
\end{equation}
Admissibility requires the historical external probability and model input to contain no information recorded after the respective forecast time. The primary model probabilities are cutoff-controlled question-only hindcasts rather than live historical model submissions. Their provenance and this limitation are described in Section~\ref{sec:experiments}.

Given a nonempty historical set, the global affine weights solve a zero-target ridge problem:
\vspace{-0.3em}
\begin{equation}
\footnotesize
\label{eq:global_weights}
\hat{\mathbf w}_{G,i}=\arg\min_{\mathbf1^\top\mathbf w=1}\Big\{\frac{1}{|\mathcal H_i|}\sum_{j\in\mathcal H_i}(\mathbf p_j^\top\mathbf w-y_j)^2+\rho_{G,i}\|\mathbf w\|_2^2\Big\}.
\end{equation}
For either historical set $\mathcal T$, we use $\rho_{\mathcal T}=10^{-3}\operatorname{tr}(\hat\Sigma_{\mathcal T})/S$, where $\hat\Sigma_{\mathcal T} =|\mathcal T|^{-1}\sum_{j\in\mathcal T}\mathbf p_j\mathbf p_j^\top$.
The penalty is centered at zero, and source probabilities are not standardized. The coefficients may be negative and represent signed error corrections, not source-reliability probabilities.

For regime $k$, set $\mathcal H_{i,k}=\{j\in\mathcal H_i:k_j=k\}$ and $n_{k,i}=|\mathcal H_{i,k}|$. When $n_{k,i}>0$, fit $\hat{\mathbf w}_{k,i}$ by replacing $\mathcal H_i$ with $\mathcal H_{i,k}$ in Eq.~\ref{eq:global_weights} and calculating its own penalty $\rho_{k,i}$ from that historical set. Both estimators use mean squared loss and the same coefficient constraint. 

\boldheader{Regime Shrinkage}
The gate interpolates between the estimates using the number of resolved examples in the current regime:
\vspace{-0.3em}
\begin{equation}
\footnotesize
\label{eq:competence_shrinkage}
\alpha_{k,i}=\frac{n_{k,i}}{n_{k,i}+\tau},\qquad \tilde{\mathbf w}_{k,i}=(1-\alpha_{k,i})\hat{\mathbf w}_{G,i}+\alpha_{k,i}\hat{\mathbf w}_{k,i}.
\end{equation}
We fix $\tau=50$ in the primary evaluation. At 50 resolved regime questions, the estimates receive equal interpolation weight. The factor $\alpha_{k,i}$ measures the amount of regime-specific history used for shrinkage, not a model's probability of being correct. The forecast contributions follow from the fitted $\tilde{\mathbf w}_{k,i}$.

\boldheader{Prediction, Calibration, and Fallback}
For question $i$, the raw affine score and probability projection are
\vspace{-0.3em}
\begin{equation}
\footnotesize
\label{eq:raw_gated_forecast}
q_i=\tilde{\mathbf w}_{k_i,i}^\top\mathbf p_i,\qquad \bar q_i=\operatorname{clip}_{[0,1]}(q_i).
\end{equation}
The returned probability is $h_i=C_i(\bar q_i)$, where $C_i$ is a chronological isotonic calibrator \cite{zadrozny2002transforming}.

The calibrator $C_i$ is fitted on historical out-of-sample predictions. For each earlier question $j$, we generate its prediction using only $\mathcal H_j$ and the same weight-fitting and fallback rules as the deployed gate. At time $t_i$, isotonic calibration uses only the resulting clipped predictions and outcomes for which $s_j<t_i$. Historical predictions are never recomputed using weights fitted after their forecast times. This prevents leakage through calibration.

When fewer than $N_{\mathrm{hist}}=150$ globally resolved examples are available, the system returns $h_i=g_i$. When global history is sufficient but the current regime has no history, it uses the global affine pool. The calibrator is the identity until $N_{\mathrm{cal}}=150$ eligible historical out-of-sample scores are available. These are fixed deployment rules, not selected using current outcomes. Clipping to $[10^{-6},1-10^{-6}]$ is used separately for logit transforms and log loss, not for the affine Brier score.

\section{Experimental Setup}
\label{sec:experiments}

\boldheader{Questions, Sources, and Forecast Provenance}
The primary development population contains 4,974 complete binary questions from nine source-defined regimes, representing 4,119 event or series clusters. The separate locked population contains 1,173 questions from 947 clusters. Markets contribute 2,941 questions from Polymarket and Manifold \cite{polymarket,manifold}. Crowds contribute 1,513 questions from INFER and ForecastBench-derived crowd questions \cite{infer,karger2025forecastbench}. Structured sources contribute 520 questions from FRED, ACLED, Wikipedia, DBnomics weather, and Yahoo Finance \cite{fred_alfred,acled,wikipedia,dbnomics,yfinance}. These groups allow us to evaluate the same deployment decision against collective forecasts and quantitative time-series priors. Source counts and the inclusion audit appear in Appendix~\ref{app:data}.

For market and crowd questions, $g_i$ is the latest recorded probability at or before $t_i$, or the explicitly supplied point-in-time probability. We never interpolate from a later snapshot. Participation measurements are likewise taken at or before $t_i$. Structured priors use historical observations available at $t_i$. For FRED, these are the latest ALFRED vintages available by that date, not subsequently revised values.

The primary language-model probabilities are generated retrospectively as \emph{cutoff-controlled question-only hindcasts}. They are not contemporaneously recorded live submissions. We require the assigned model knowledge cutoff to precede the question's forecast date by at least 31 days and exclude post-$t_i$ evidence from its input. This reduces known forms of leakage but cannot prove that model pretraining lacked all relevant outcome information. The registry comprises 6,269 records: 4,974 development questions, 1,173 locked questions, 120 gap questions, and two invalid primary-panel parses. The row-level provenance manifest and filtering rules appear in Appendix~\ref{app:data}.

\boldheader{Model Panel and Elicitation}
The primary panel contains gpt-oss-20b, gpt-oss-120b, Llama-3.1-8B-Instruct, and Llama-3.3-70B-Instruct. These models allow us to examine relative competence across two families and multiple model sizes. Each receives the question, forecast date, and resolution criteria without the external probability or retrieval. Primary predictions use temperature zero and are parsed into one probability. Complete elicitation settings appear in Appendix~\ref{app:models}. 

\boldheader{External Forecasts and Calibration}
For structured sources, $g_i^{\mathrm{TS}}$ is the equal-weight mean of valid event probabilities from persistence, ETS, ARIMA, and a local-level state-space model. ACLED adds a Poisson or overdispersed negative-binomial count component. At least two components must be valid. Otherwise, the prior is the same-horizon empirical event probability with Jeffreys smoothing. This fallback is used on 20 of 520 structured questions. The conversion of predictive distributions into binary event probabilities is given in Appendix~\ref{app:priors}.

We compare against a chronological external-only Platt reference learned from the same eligible resolved history without LM inputs. The gate receives the raw external probability, while the calibrated reference controls for improvements achievable without combining sources. An external-only isotonic comparator, evaluated on the same 4,974 questions, obtains 0.07556 Brier versus 0.07573 for Platt calibration. Appendix~\ref{app:baselines} gives both calibration procedures and the fallback rules.

\boldheader{Baselines and Matched Controls}
We evaluate individual models, an LLM-only pool, global combinations, and regime-specific combinations. Matched simplex and signed affine comparisons hold source inputs and calibration fixed while varying the coefficient constraint and whether weights depend on regime. The calibrated global affine pool isolates the contribution of routing, while the calibrated external-only baseline tests incremental hybrid value.

We also evaluate Local Prediction Pools, regime-aware logistic stacking, hierarchical logistic stacking, and an XGBoost router. These alternatives use the same forecast-time information and provide more flexible aggregation rules. Their features, regularization, and chronological fitting procedures are specified in Appendix~\ref{app:baselines}. They serve as accuracy controls rather than methods the simpler gate is assumed to outperform.

\boldheader{Chronological Protocol, Hyperparameters, and Tests}
We replay predictions in forecast-time order. An outcome becomes eligible for fitting only after its recorded resolution-availability time. The primary gate uses the fixed shrinkage and ridge settings defined in Section~\ref{sec:method}. We vary $\tau$ and the ridge multiplier only in sensitivity analyses, beginning after at least 300 resolved historical examples. The sensitivity grid and numerical safeguards are reported in Appendix~\ref{app:hyperparameters}.

The primary endpoint is the mean paired Brier improvement over calibrated external forecasts:
\begin{equation}
\label{eq:primary_gain}
\Delta_1=\widehat{\mathcal B}(g^{\mathrm{cal}})
-\widehat{\mathcal B}(h_{\mathrm{gate}}),
\end{equation}
where $\widehat{\mathcal B}$ denotes the empirical mean Brier loss on the evaluation population. We also report Brier skill,
$1-\widehat{\mathcal B}(h_{\mathrm{gate}})/
\widehat{\mathcal B}(g^{\mathrm{cal}})$.
The secondary routing endpoint is
\begin{equation}
\label{eq:empirical_routing_gain}
\Delta_2=\widehat{\mathcal B}(h_G)
-\widehat{\mathcal B}(h_{\mathrm{gate}}),
\end{equation}
where $h_G$ uses global affine weights with matched isotonic calibration. The separate raw routing diagnostic compares unprojected, uncalibrated global and regime affine scores.

We report paired event-clustered uncertainty with Wald intervals. Full inference procedures and power checks appear in Appendix~\ref{app:robustness}. Routing is a secondary endpoint, and its confidence interval includes gains smaller than the prespecified 0.001 Brier threshold of practical interest.

Development ends December 31, 2025. We exclude January 1 through February 25, 2026 to separate development from locked evaluation, while event-disjoint splits prevent overlap in underlying targets. Locked evaluation begins February 26, 2026 and is used as a consistency check, not for method selection.

\section{Results}
\label{sec:results}

\boldheader{Weaker Standalone Models Improve the External Forecast}
On the 4,974-question development population, the best standalone LM obtains 0.09461 Brier, 24.94\% higher loss than the calibrated external forecast at 0.07573. The optimized LM-only pool also performs worse than the external reference. Standalone superiority is therefore not necessary for complementary value.

The competence gate reduces Brier to 0.07319, a gain of 0.00254 or 3.35\% Brier skill (95\% CI for the gain [0.00136, 0.00371]; $p=2.24\times10^{-5}$). The external-only isotonic comparator obtains 0.07556 Brier, so the gain does not vanish under matched calibration families. Table~\ref{tab:main_results} reports the matched comparisons.

Flexible contextual aggregators achieve slightly lower Brier point estimates, but the reported paired comparisons do not establish a significant difference from the gate. The gate therefore offers an interpretable source-level combination rather than a demonstrated accuracy advantage over flexible stacking. Its fitted coefficients make the historical corrections attributed to each source explicit.


\begin{table}[t]
\centering
\scriptsize
\setlength{\tabcolsep}{3.5pt}
\setlength{\aboverulesep}{0pt}
\setlength{\belowrulesep}{0pt}
\renewcommand{\arraystretch}{1.05}
\begin{tabular}{lrr}
\toprule
\rowcolor{headerblue}
\textbf{Method} & \textbf{Brier} $\downarrow$ & \textbf{BSS (\%)} $\uparrow$ \\
\midrule
\multicolumn{3}{l}{\scriptsize\textit{External and LM-only}} \\
Calibrated external (reference) & 0.07573 & 0.00 \\
Best standalone LM              & 0.09461 & $-24.94$ \\
Optimized LM-only pool          & 0.08731 & $-15.30$ \\
\midrule
\multicolumn{3}{l}{\scriptsize\textit{Matched hybrid pools}} \\
Global affine + isotonic        & 0.07428 & 1.91 \\
Regime simplex + isotonic       & 0.07386 & 2.47 \\
\textbf{Competence gate (ours)} & \textbf{0.07319} & \textbf{3.35} \\
\midrule
\multicolumn{3}{l}{\scriptsize\textit{Flexible aggregators}} \\
Local Prediction Pools          & 0.07316 & 3.40 \\
Hierarchical stacker            & 0.07308 & 3.49 \\
Regime-aware logistic stacker   & 0.07306 & 3.52 \\
Contextual XGBoost              & \underline{0.07302} & \underline{3.57} \\
\bottomrule
\end{tabular}
\caption{Brier score and Brier skill score (BSS) on the common 4,974-question population. BSS is measured relative to the chronologically calibrated external forecast. \textbf{Bold}: ours; \underline{underlined}: lowest Brier. The gate improves the reference by 3.35\% (95\% CI 1.80--4.90\%); differences from flexible aggregators are not statistically significant.}
\label{tab:main_results}
\end{table}
\boldheader{Regime Conditioning Contributes Beyond Global Pooling}
We isolate regime conditioning by comparing global and regime-specific pools with identical inputs, coefficient constraints, and calibration. Table~\ref{tab:routing_ablation} shows positive regime-specific gains under all three matched configurations. The raw affine improvement is 0.00104 Brier before clipping and recalibration. With matched isotonic calibration, the routing gain is 0.00109 Brier, with a 95\% confidence interval of [0.00033, 0.00186] and paired $p=0.00504$.

Routing explains only part of the full hybrid improvement. The calibrated regime simplex obtains 0.07386 Brier, compared with 0.07319 for the signed affine gate, indicating an additional contribution from the coefficient constraint. Recalibration also improves the final pool. The routing result is secondary because its confidence interval includes gains below the prespecified 0.001 Brier threshold of practical interest.


\begin{table}[t]
\centering
\small
\setlength{\tabcolsep}{3.7pt}
\setlength{\aboverulesep}{0pt}
\setlength{\belowrulesep}{0pt}
\begin{tabular}{lrrr}
\toprule
\rowcolor{headerblue}
\textbf{Pool} & \textbf{Global} & \textbf{Regime} & \textbf{Gain} $\uparrow$ \\
\midrule
Affine, raw       & 0.07510 & 0.07406 & 0.00104 \\
Simplex, isotonic & 0.07475 & 0.07386 & 0.00089 \\
\textbf{Affine, isotonic} & 0.07428 & \textbf{0.07319} & \textbf{0.00109} \\
\bottomrule
\end{tabular}
\caption{Matched global and regime-specific pooling comparisons (Brier). Gain is
global minus regime-specific Brier; positive values favor regime-specific pooling.
The raw affine row uses scores before clipping and recalibration.
\textbf{Bold}: the deployed gate configuration.}
\label{tab:routing_ablation}
\end{table}

\boldheader{Complementary Value Depends on the External Source}
Complementary value varies across the nine forecasting regimes. The gate improves Manifold by approximately 0.00371 Brier, the ForecastBench-derived crowd subset by 0.00305, and INFER by 0.00477. In contrast, the Polymarket gain is approximately 0.00038, with a confidence interval spanning zero. The smaller correction is consistent with limited additional model information in this market setting, although it does not establish that every Polymarket forecast is stronger.

Structured questions provide a separate test against quantitative time-series priors. Across 520 questions, the calibrated classical ensemble obtains 0.17374 Brier, compared with 0.16909 for the hybrid, a gain of 0.00465. FRED has a positive source-level improvement, while the smaller structured groups have less precise estimates. DBnomics weather and Yahoo Finance show near-zero gains with intervals spanning zero. These results indicate that the gate's value depends on the competing source and is not uniformly positive across forecasting tasks.

\begin{table}[t]
\centering
\small
\setlength{\tabcolsep}{3.2pt}
\setlength{\aboverulesep}{0pt}
\setlength{\belowrulesep}{0pt}
\begin{tabular}{lrrr}
\toprule
\rowcolor{headerblue}
\textbf{Source} & \textbf{$n$} & \textbf{External} & \textbf{Gate} \\
\midrule
Polymarket          & 1,904 & 0.06008 & 0.05971 \\
Manifold            & 1,037 & 0.07310 & \textbf{0.06939} \\
ForecastBench crowd &   912 & 0.05820 & \textbf{0.05515} \\
INFER               &   601 & 0.07160 & \textbf{0.06683} \\
FRED                &   214 & 0.16784 & \textbf{0.15995} \\
ACLED               &    72 & 0.15493 & 0.14907 \\
Wikipedia           &    61 & 0.14087 & 0.13651 \\
DBnomics weather    &    78 & 0.15008 & 0.14996 \\
Yahoo Finance       &    95 & 0.24181 & 0.24149 \\
\bottomrule
\end{tabular}
\caption{Brier scores by external-source regime. \textbf{Bold}: gate improvement
whose event-clustered 95\% confidence interval excludes zero. The ForecastBench crowd
row identifies the dataset distribution source while retaining the underlying crowd
regime. Counts sum to 4,974. Confidence intervals appear in
Appendix~\ref{app:regimes}.}
\label{tab:regime_results}
\end{table}

\boldheader{Model Contribution Falls with Market and Crowd Participation}
We examine whether participation is associated with the correction assigned to LM forecasts. Market volume and trader count, or the number of crowd forecasters, provide forecast-time participation proxies. After ranking these measurements within each source, we form four equal-count bins from 4,454 market and crowd questions.

\begin{figure}[t]
\centering
\includegraphics[width=\columnwidth]{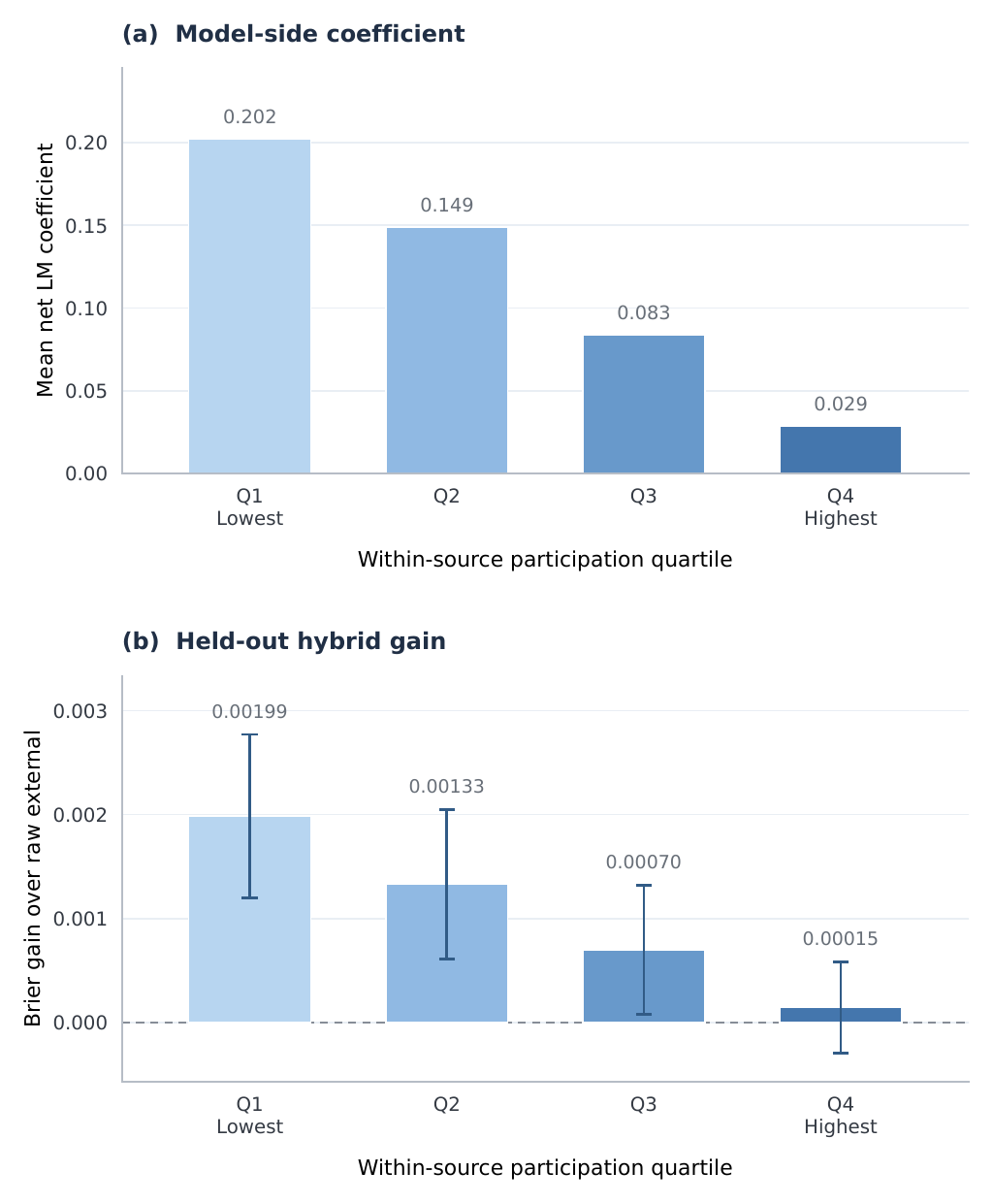}
\caption{Participation and LM correction across 4,454 market and crowd questions. (a) Mean net signed LM coefficient by within-source participation quartile. (b) Brier gains over the raw external forecast from a separate scalar diagnostic, evaluated on the later 20\% of questions within each source-quartile. Error bars show 95\% confidence intervals, and the horizontal line marks zero gain. Participation is not a direct measure of accuracy, and the association is observational.}
\label{fig:external_strength}
\end{figure}

Because the affine source weights sum to one, the net model-side coefficient for question $i$ is
\begin{equation}
\label{eq:effective_model_coefficient}
\lambda_i^{\mathrm{eff}}
=
\sum_{m=1}^{M}\tilde w_{k_i,i,m}
=
1-\tilde w_{k_i,i,g}.
\end{equation}
Here $\tilde w_{k_i,i,g}$ denotes the external-source coefficient, and $\tilde w_{k_i,i,m}$ denotes the coefficient of model $m$. Mean net model-side affine correction coefficients decrease from 0.20179 in the lowest-participation quartile to 0.02860 in the highest, a difference of 0.17319 (95\% CI, [0.09820, 0.24819]). A separate scalar later-held-out participation diagnostic shows declining Brier gains from 0.00199 to 0.00015 across these bins. Its weighted mean gain is 0.00104, distinct from the 0.00229 gain of the full multi-model gate on the market and crowd population. Figure~\ref{fig:external_strength} shows both trends.


The negative association also appears in a regression controlling for source identity, forecast horizon, and external probability extremity. Higher-participation markets and larger crowds thus receive smaller model-side corrections in the evaluated sources. This is consistent with broader external aggregation leaving less complementary information, but participation is not a direct measure of accuracy and the relationship is observational.

\boldheader{Historical Confidence as a Secondary Diagnostic}
Historical comparative competence predicts individual-model wins over the raw external forecast with model--question-pair AUC 0.681, compared with raw verbal-confidence AUCs of 0.463--0.501. It also improves question-level selection of hybrid benefit, with relative AURC $-0.00502$ versus $-0.00295$ for history-oriented verbal confidence. This diagnostic does not feed the gate; definitions and comparisons appear in Appendix~\ref{app:confidence}.

\boldheader{Temporal and Mechanism Checks}
The separate locked evaluation shows a positive but inconclusive gain of 0.00227 Brier (95\% CI, $[-0.00040,0.00494]$) across 1,173 questions. Its interval includes zero. Raw routing and sensitivity checks are reported in Appendices~\ref{app:theory}, \ref{app:hyperparameters}, and \ref{app:robustness}.

\section{Discussion and Conclusion}
\label{sec:discussion}
\label{sec:conclusion}

Our findings show that forecasting competence depends on the information already available. Under Brier loss, a weaker standalone LM can improve an external forecast when its disagreement systematically corrects external errors. The competence gate learns inspectable source-specific corrections from resolved outcomes and improves a calibrated external forecast while achieving accuracy comparable to more flexible aggregators. Model-side corrections decrease with market and crowd participation, although this relationship is observational. These findings support evaluating LMs by their complementary value rather than standalone accuracy or self-reported confidence. Hybrid systems should preserve strong external forecasts when historical evidence does not support additional corrections. Further discussion appears in Appendix~\ref{app:discussion}.

Our results shift the deployment question from whether an LM is a better forecaster to whether it contributes information beyond an existing prediction. Relative competence provides a way to characterize this value, while historical source-specific corrections offer a practical way to estimate it. The broader implication is that hybrid forecasting systems should adapt model influence to the competing information source rather than assign uniform trust to LMs.

\section{Limitations}
\label{sec:limitations}

Our evaluation covers resolved binary questions from nine forecasting sources. Generalization to continuous outcomes, other information sources, and different event distributions remains untested. Smaller source populations limit estimation precision, and shrinkage cannot guarantee adaptation to changing source quality or capture finer question-level relationships. The primary LM forecasts are cutoff-controlled retrospective hindcasts rather than live predictions. Historical eligibility restrictions reduce identifiable leakage but cannot rule out training-data contamination or errors in information availability. The routing gain remains secondary, and the locked evaluation is insufficiently powered for independent confirmation. Participation associations are observational, and the theoretical routing identity does not guarantee finite-sample improvement. Finally, Brier-score gains do not establish better downstream decisions or account for inference costs, latency, or consequences of errors. Prospective evaluation is needed.

\newpage


\bibliography{custom}

\appendix

\section*{Impact Statement}

Probabilistic forecasts support decisions in areas such as public policy, resource planning, finance, and disaster response. This work studies how language-model predictions can be combined with forecasts that are already available from other sources. Estimating complementary predictive value may improve forecast aggregation and help practitioners identify settings where additional model predictions contribute limited information. The proposed method also provides inspectable source coefficients that can support analysis of historical forecasting behavior.

Forecasting errors can have consequential effects, and historical competence does not guarantee future reliability. Source quality may change, model predictions may reflect incomplete or contaminated information, and evaluation datasets may not represent the conditions of a particular deployment. The reported improvements concern probabilistic forecasting accuracy, not downstream decision quality. Applications involving consequential decisions require independent validation, ongoing monitoring, and appropriate human oversight.

\section{Additional Discussion and Implications}
\label{app:discussion}

\paragraph{Forecasting value is relative to the available information.}
Standalone accuracy is not sufficient to determine whether a language model should be used in a hybrid forecasting system. A model can be less accurate than an external source yet provide information that improves the combined forecast. Our results demonstrate this distinction across markets, forecasting crowds, and structured prediction tasks. The Brier-loss analysis identifies the underlying condition: model disagreement is useful when it systematically corrects errors in the external forecast. This shifts the deployment question from whether a model is the better forecaster to whether it contributes information beyond what is already available. The distinction also applies to forecast combinations involving statistical models and human judgments, where the value of an additional source depends on the information represented by the existing ones.

\paragraph{Interpretable combination is useful even without superior accuracy.}
The competence gate provides a simple way to estimate how source contributions differ across forecasting regimes. Its coefficients are learned from resolved outcomes, with shrinkage limiting the influence of sparse regime histories. The matched experiments show an additional benefit from regime conditioning, although this contribution is smaller and less firmly established than the overall hybrid gain. More flexible contextual aggregators achieve similar predictive accuracy. The gate's contribution is therefore not a claim of universal superiority over stacking methods. Beyond conventional pooling, we study source-relative predictive value, characterize it under Brier loss, and test source-conditioned corrections using matched controls. The fitted coefficients make those corrections inspectable. This makes the relationship between observed forecasting errors and the resulting source weights easier to examine.

\paragraph{Complementary value changes with the competing forecast.}
The measured benefit of language-model forecasts varies across external sources. Improvements are small in the evaluated Polymarket and near-zero-gain structured regimes. Within market and crowd sources, higher participation is associated with smaller model-side corrections, consistent with less complementary information remaining after external aggregation. This is an observational relationship, not evidence that participation causes greater forecast accuracy. The confidence results reinforce the same distinction between standalone performance and complementary value. Historical comparative signals are more informative than raw verbal confidence, but predicting when an individual model wins does not necessarily identify when the hybrid improves. Together, these findings support estimating model contributions from resolved outcomes and the external forecasting context rather than treating language-model confidence or standalone rankings as sufficient deployment criteria.

\section{Population Construction and Point-in-Time Provenance}
\label{app:data}
The registry contains 6,269 records: 4,974 development complete cases, 1,173 questions in the separate locked population, 120 questions in the eight-week gap, and two invalid primary-panel parses. The development rows belong to 4,119 event or series clusters. The locked population contains 947 event-disjoint clusters. Source counts are 1,904 Polymarket, 1,037 Manifold, 912 ForecastBench-derived crowd, 601 INFER, 214 FRED, 72 ACLED, 61 Wikipedia, 78 DBnomics weather, and 95 Yahoo Finance. A given event or series contributes one resampling cluster even when it has several forecast snapshots.

The canonical row-level manifest records \texttt{question\_id}, \texttt{event\_id}, \texttt{forecast\_time}, \texttt{resolution\_available\_time}, regime, snapshot timestamps and identifiers, ALFRED vintage dates, model-cutoff eligibility, raw probabilities, parsing-validity flags, split, forecast artifact hashes, and event-cluster membership. The model record additionally preserves \texttt{model\_id}, question-text hash, prompt hash, and generation commit. Source values and participation proxies must be from at or before $t_i$. These fields determine every analysis population and support an audit of information available at the question timestamp. Because the LLM predictions are retrospectively generated hindcasts, cutoff eligibility is a contamination control rather than evidence that the forecasts were recorded live.

\paragraph{ForecastBench provenance.}
The 912-question crowd subset was obtained from the publicly released ForecastBench dataset \cite{karger2025forecastbench}, rather than collected through the original forecasting platform's API. We retain the underlying source identity for regime construction and historical aggregation. The ForecastBench dataset repository distributes its data under CC BY-SA 4.0. Dataset attribution and the frozen release identifier are retained in the provenance manifest.

\section{Model Input and Reproducibility}
\label{app:models}
The primary panel has two gpt-oss and two Llama instruct models. The frozen cutoff assignment is June 1, 2024 for gpt-oss. For the month-level December 2023 cutoff reported for Llama 3.1 and 3.3, we conservatively use December 31, 2023 when applying the 31-day eligibility buffer. Every eligible question satisfies the 31-day buffer. The input contains only question text, resolution criteria, and forecast date, with no external probability, retrieval, or post-date evidence. The primary output is one parsed probability at temperature zero. The gpt-oss models use medium reasoning effort and a 4,096-token output limit. The Llama models use their native chat templates and a 1,024-token output limit.

For reproducibility, the release includes checkpoint identifiers, input templates, the probability parser, inference software versions, generation hashes, and model-specific hardware. The event-level manifest links each parsed prediction to its model, prompt, and forecast date. Because the primary panel uses question-only models without retrieval, these experiments do not establish whether the same gains hold for retrieval-augmented or more capable frontier models. The external-conditioned elicitation control in Appendix~\ref{app:additional_controls} tests access to the external probability, not retrieval augmentation.

\section{Quantitative External Prior}
\label{app:priors}
For FRED, Wikipedia, DBnomics, and Yahoo Finance, eligible probabilities come from persistence, ETS, ARIMA, and local-level state space. ACLED additionally uses a Poisson or negative-binomial count model, selecting negative binomial under overdispersion. If $\mathcal J_i$ indexes the valid components and $|\mathcal J_i|\ge2$, their equal-probability ensemble is
\begin{equation}
\label{eq:ts_ensemble}
g_i^{\mathrm{TS}}=\frac1{|\mathcal J_i|}
\sum_{m\in\mathcal J_i}g_{im}^{\mathrm{TS}}.
\end{equation}
When all components are valid, there are four equal weights except for ACLED, which has five.

For a continuous forecast distribution with CDF $F$, $\Pr(X>c)=1-F(c)$ and $\Pr(X\le c)=F(c)$. For integer counts, $\Pr(X\ge c)=1-F(c-1)$ and $\Pr(X>c)=1-F(c)$. The threshold direction and forecast horizon follow the question. We average event probabilities, not point estimates. For FRED, the estimation sample is the latest ALFRED vintage available at $t_i$, rather than modern revisions of earlier observations.

If fewer than two components are valid, the fallback uses the historical same-horizon frequency with Jeffreys smoothing:
\begin{equation}
\label{eq:ts_fallback}
g_i^{\mathrm{TS}}=\frac{x+1/2}{n+1},
\end{equation}
where $x$ is the number of event occurrences among $n$ available past comparisons. This case occurs on 20 of 520 structured questions. Classical model fitting and predictive-distribution construction must obey the historical cutoff. Component weights remain fixed and equal and are not selected using forecasting-question outcomes. Time-series foundation-model robustness results appear in Table~\ref{tab:foundation_models} and are not inputs to $g_i^{\mathrm{TS}}$.

\section{Additional Quantitative and Information-Set Controls}
\label{app:additional_controls}
\subsection{Time-Series Foundation-Model Comparisons}
The classical structured reference is compared with three time-series foundation models on the structured evaluation. These foundation models are robustness references rather than members of the primary equal-weight prior. The hybrid using the classical prior obtains 0.169090 Brier, versus 0.173740 for the calibrated classical prior. Table~\ref{tab:foundation_models} reports the additional comparisons.
\begin{table}[t]
\centering\small
\begin{tabular}{lr}
\hline
\rowcolor{blue!10}
Method & Brier $\downarrow$\\
\hline
Calibrated classical prior & 0.173740\\
Chronos & 0.175811\\
TimesFM & 0.176504\\
Moirai & 0.176123\\
\textbf{Hybrid with classical prior} & \textbf{0.169090}\\
\hline
\end{tabular}
\caption{Structured-event forecasting on 520 questions. Foundation-model predictions are separate reference forecasts and are not components of the primary classical prior.}
\label{tab:foundation_models}
\end{table}

\subsection{Independent and External-Conditioned Elicitation}
We compare the primary question-only LLM forecasts with an elicitation variant that shows the model the external probability. The measured combination value decreases from 0.001532 to 0.000718, a 53.1\% reduction. The standalone Brier reported for this matched elicitation comparison improves from 0.101694 to 0.097881; it is a separate summary from the best-single-model result in the main results table. The correlation between model and external probabilities rises from 0.603814 to 0.841117, while mean absolute disagreement decreases from 0.126441 to 0.071932. The net model-side affine coefficient decreases from 0.132814 to 0.063184. The measurements are consistent with increased imitation of the external signal after exposure, but do not establish that imitation alone caused the loss of complementary value.
\begin{table}[t]
\centering\small
\setlength{\tabcolsep}{4pt}
\begin{tabular}{lrr}
\hline
\rowcolor{blue!10}
Metric & Independent & Conditioned\\
\hline
Standalone Brier (elicitation comparison) $\downarrow$ & 0.101694 & 0.097881\\
Combination value $\uparrow$ & 0.001532 & 0.000718\\
Corr. $(f,g)$ & 0.603814 & 0.841117\\
Mean $|f-g|$ & 0.126441 & 0.071932\\
Model-side coefficient & 0.132814 & 0.063184\\
\hline
\end{tabular}
\caption{Independent question-only elicitation versus forecasts generated after revealing the external probability. Both variants are evaluated on the same 4,974 development questions. The conditioned forecast is a separate elicitation variant. The primary gate uses independent question-only forecasts.}
\label{tab:conditioned_elicitation}
\end{table}

\subsection{Regime Conditioning, Clipping, Shrinkage, and Calibration}
Matched ablations isolate the effects of regime conditioning, clipping, shrinkage, and calibration. Table~\ref{tab:calibration_ablations} reports distinct controlled comparisons, not consecutive operations in one prediction pipeline. The deployed method shrinks fitted weights before pooling, clipping, and chronological isotonic calibration.

\begin{table}[t]
\centering\small
\setlength{\tabcolsep}{4pt}
\begin{tabular}{lr}
\hline
\rowcolor{blue!10}
Control & Brier $\downarrow$\\
\hline
Raw global affine & 0.075101842\\
Raw regime affine & 0.074064247\\
Clipped regime affine & 0.074028511\\
Shrinkage gate, before calibration & 0.073842610\\
Final calibrated gate & 0.073187332\\
\hline
\end{tabular}
\caption{Matched ablation controls on the development population. Each row reports its stated estimator. The rows are not sequential deployment stages.}
\label{tab:calibration_ablations}
\end{table}

Relative to raw global affine pooling, the raw regime affine control improves Brier by 0.001037595. Clipping the regime affine score changes Brier by 0.000035736. The pre-calibration shrinkage-gate control differs from the clipped regime affine control by 0.000185901, and chronological calibration changes the shrinkage-gate score by 0.000655278. These are matched control contrasts, not consecutive deployment stages.

\section{Baseline Implementations and Calibration}
\label{app:baselines}
\subsection{Matched Linear Pools}
The global and regime-specific affine models use the same historical forecasts, mean squared-error objective, zero-centered ridge penalty, and sum-to-one coefficient constraint. A simplex pool additionally imposes $w_j\geq0$ on every source. Matched raw and calibrated comparisons use identical evaluation questions. Their calibration maps are fitted chronologically, with each historical pooled prediction generated using only the information available at its own forecast date.

\subsection{External-Only Platt Calibration}
The external-only reference is learned without LLM inputs. For each historical or test probability, let
\begin{equation}
\label{eq:app_external_logit}
u_j=\operatorname{logit}\!\left(
\operatorname{clip}_{[10^{-6},1-10^{-6}]}(g_j)\right).
\end{equation}
At time $t_i$, global Platt parameters $(a_{G,i},b_{G,i})$ minimize the mean binary logistic loss over $\mathcal H_i$, with an additional penalty $10^{-4}(a^2+b^2)$. When regime $k$ has eligible outcomes, the corresponding parameters $(a_{k,i},b_{k,i})$ are fitted using $\mathcal H_{i,k}$ and the same objective. We shrink the parameter vectors, not their output probabilities, using $\beta_{k,i}=n_{k,i}/(n_{k,i}+50)$ and $\tilde{\boldsymbol\theta}_{k,i}=(1-\beta_{k,i})\hat{\boldsymbol\theta}_{G,i}+\beta_{k,i}\hat{\boldsymbol\theta}_{k,i}$, where $\boldsymbol\theta=(a,b)^\top$. The final comparator is $\sigma(\tilde a_{k,i}+\tilde b_{k,i}u_i)$. A regime with no eligible examples uses the global fit. If fewer than 150 globally resolved outcomes are available, the comparator returns $g_i$. The global and regime Platt fits use the same chronologically eligible historical question sets $\mathcal H_i$ and $\mathcal H_{i,k}$ as the primary analysis, but use only $g$ and $y$. The Platt penalty is independent of the gate's ridge penalty.

\paragraph{External-only isotonic control.}
On the matched 4,974-question development population, chronological external-only isotonic calibration obtains 0.075560 Brier, versus 0.075725674 for the primary external-only Platt comparator and 0.073187332 for the gate. The paired uncertainty reported for the Platt comparison does not apply to this additional control.

\subsection{Contextual Comparators}
We construct contextual features from information available at the forecast time. Let $T_i$ be the forecast horizon in days, $\bar f_i=M^{-1}\sum_m f_{im}$, and let $s_f(i)$ be the sample standard deviation of the $M$ model probabilities. Define $D_i=M^{-1}\sum_m|f_{im}-g_i|$, $E_i=|g_i-0.5|$, and $\ell(p)=\operatorname{logit}(\operatorname{clip}_{[10^{-6},1-10^{-6}]}(p))$. These transformations provide source probabilities, cross-model dispersion, external extremity, and disagreement without using outcomes or later information.

\paragraph{Regime-aware logistic stacking.}
We use the 14 continuous features
\begin{equation}
\label{eq:logistic_features}
\begin{aligned}
\mathbf v_i=[&\ell(g_i),\ell(f_{i1}),\ldots,\ell(f_{i4}),\\
&f_{i1}-g_i,\ldots,f_{i4}-g_i,\\
&\bar f_i,s_f(i),D_i,E_i,\log(1+T_i)]^\top.
\end{aligned}
\end{equation}
The predictor is $\sigma(\beta_0+\beta^\top\mathbf v_i+\gamma_{k_i,0})$, with global feature coefficients and regime-specific intercept indicators. We standardize the continuous features using the eligible historical mean and standard deviation and leave regime indicators unscaled. We fit L2 logistic regression using \texttt{lbfgs}, an intercept, no class weighting, \texttt{max\_iter=5000}, and \texttt{tol=1e-8}. We choose $C\in\{0.01,0.1,1,10\}$ on the most recent 20\% of chronologically eligible history and refit on all eligible history with the selected value. The ordinary logistic stack has no regime-specific feature slopes, leaving that flexibility to the hierarchical comparator.

\paragraph{Hierarchical logistic stacking.}
We use the same 14 continuous features but allow regime-dependent intercepts and slopes:
\begin{equation}
\label{eq:hierarchical_stacker}
\eta_i=\beta_0+\beta^\top\mathbf v_i+\gamma_{k_i,0}
+\gamma_{k_i}^\top\mathbf v_i.
\end{equation}
The global L2 penalty is fixed at 1, while the regime-deviation penalty is selected chronologically from $\{1,10,100\}$. This is a partial-pooling comparator that can represent within-regime relationships absent from the proposed source-level affine gate.

\paragraph{Contextual gradient boosting.}
The XGBoost comparator uses the same forecast-time contextual features and regime information as logistic stacking. It uses \texttt{objective=binary:logistic}, learning rate 0.03, maximum depth 3, minimum child weight 20, row and column subsampling of 0.8, L2 regularization 1, and at most 500 trees. The most recent 20\% of eligible historical examples are reserved for chronological early stopping with patience 30. We then refit on all eligible history using the selected number of trees.

\paragraph{Local Prediction Pools.}
We implement the caliper Local Prediction Pool of \citet{oelrich2024local}, using six forecast-time pooling variables:
\begin{equation}
\label{eq:lpp_features}
\begin{aligned}
\mathbf z_i=[&\ell(g_i),E_i,\log(1+T_i),\\
&\bar f_i,s_f(i),D_i]^\top.
\end{aligned}
\end{equation}
We standardize each pooling coordinate using eligible historical means and standard deviations only. Neighbors come from the current regime's historical set $\mathcal H_{i,k_i}$, with Euclidean distance $d_{ij}=\|\tilde{\mathbf z}_i-\tilde{\mathbf z}_j\|_2$. For caliper width $\varrho$, the local set is $\mathcal N_i(\varrho)=\{j\in\mathcal H_{i,k_i}:d_{ij}\leq\varrho\}$. We choose $\varrho\in\{0.50,0.75,1.00,1.50,2.00,3.00\}$ on the most recent 20\% of available historical observations, respecting chronology, and recompute local weights using all earlier resolved history at the selected width.

For source $s\in\{0,\ldots,M\}$ with probability $p_{js}$, its local mean Bernoulli log score is
\begin{equation}
\label{eq:lpp_local_ability}
\hat a_{is}=
\frac{1}{|\mathcal N_i|}\sum_{j\in\mathcal N_i}
\left[y_j\log p_{js}^{\epsilon}
+(1-y_j)\log(1-p_{js}^{\epsilon})\right],
\end{equation}
where $p_{js}^{\epsilon}=\operatorname{clip}_{[10^{-6},1-10^{-6}]}(p_{js})$ and the formula applies when $|\mathcal N_i|>0$. Following the natural-scaling version of the published caliper pool, weights depend on the local \emph{sum} of log scores, rather than an unscaled mean:
\begin{equation}
\label{eq:lpp_weights}
\omega_{is}=
\frac{\exp(|\mathcal N_i|\hat a_{is})}
{\sum_{r=0}^{M}\exp(|\mathcal N_i|\hat a_{ir})},
\quad
h_i^{\mathrm{LPP}}=\sum_{s=0}^{M}\omega_{is}p_{is}.
\end{equation}
If the caliper contains no observations, we set all source weights to $1/S$. The predicted probability is a convex pool in $[0,1]$. The $10^{-6}$ clip is used only in the log-score calculation. Local pooling uses the same historical eligibility and observed regime partition as the gate, but learns question-dependent rather than regime-constant source weights.

\subsection{Forecast-Information Parity}
All contextual methods are trained on chronological, resolved history and receive only forecast-time features. They do not receive future market activity, future crowd sizes, later time-series observations, resolved outcomes for current questions, or realized gate errors. The contextual alternatives achieve Brier scores of 0.073061 (logistic), 0.073082 (hierarchical logistic), 0.073155 (Local Prediction Pools), and 0.073021 (XGBoost), compared with 0.073187 for the gate. None has a statistically established advantage over the gate in the reported paired comparisons. These results indicate comparable predictive accuracy, not superiority of the gate.

\section{Sensitivity and Numerical Stability}
\label{app:hyperparameters}
The primary ridge multiplier is fixed at $10^{-3}$ and $\tau$ at 50. Sensitivity analyses evaluate $\tau\in\{10,25,50,100,200\}$ and
\begin{equation}
\label{eq:ridge_grid}
c_\rho\in\{2.5\!\times\!10^{-4},5\!\times\!10^{-4},
10^{-3},2\!\times\!10^{-3},4\!\times\!10^{-3}\}.
\end{equation}
The sensitivity analysis is evaluated once at least 300 historical cases are available and does not retune primary predictions. The second moment is uncentered and calculated separately for each eligible global or regime fit. Mean squared error is used throughout. A summed-loss ridge solver must use $n\rho$. The condition-number audit threshold is $10^5$ with family-mean stabilization when required. The audit records activation of the $10^5$ condition-number guard separately from the ordinary fitted-weight distribution.

In the fitted-weight audit, 21.47\% of affine coefficients are negative, and 1.99\% of raw affine outputs fall outside $[0,1]$. The latter are clipped for probability evaluation but not in the raw theory diagnostic. The best evaluated shrinkage--ridge grid setting obtains 0.073083 Brier, compared with 0.073187 for the frozen primary configuration, a difference of 0.000104. The regime-aware logistic stacker obtains 0.073061 Brier and is not part of this sensitivity grid. A separate uncertainty-aware variant reduces the negative-transfer rate from 0.155227 to 0.143884, a 7.31\% relative reduction, with nearly unchanged Brier loss. This variant is not the primary estimator.

\section{Population Routing Proof and Mechanism Controls}
\label{app:theory}
For any regime with $A_k>0$, completing the square in Eq.~\ref{eq:affine_risk} gives
\begin{equation}
\label{eq:completed_square}
\mathcal R_k(\lambda)=\mathcal R_k(\lambda_k^\star)
+A_k(\lambda-\lambda_k^\star)^2.
\end{equation}
Minimizing the weighted sum over a shared $\lambda$ yields Eq.~\ref{eq:optimal_global_weight}. Substituting the global optimum into Eq.~\ref{eq:completed_square} and subtracting the regime-specific minima proves Eq.~\ref{eq:routing_identity}.

For $S$ sources with weights on the affine hyperplane, choose any matrix $U \in \mathbb{R}^{S \times (S-1)}$ whose columns span the zero-sum subspace and any feasible $\mathbf{w}_{\mathrm{ref}}$ with $\mathbf{1}^\top \mathbf{w}_{\mathrm{ref}} = 1$. Write $\mathbf{w} = \mathbf{w}_{\mathrm{ref}} + U \boldsymbol{\theta}$. Define the projected source feature $\mathbf{a} = U^\top \mathbf{p}$, residual $e_{\mathrm{ref}} = \mathbf{p}^\top \mathbf{w}_{\mathrm{ref}} - y$, and $H_k = \mathbb{E}_k[\mathbf{a} \mathbf{a}^\top]$. Assuming $H_k$ is positive definite on this subspace, the regime optimizer is $\boldsymbol{\theta}_k^\star = -H_k^{-1} \mathbb{E}_k[\mathbf{a} e_{\mathrm{ref}}]$. The multi-source population routing gain is
\begin{equation}
\label{eq:multisource_identity}
\begin{aligned}
&\sum_k\pi_k\mathcal R_k(\boldsymbol\theta_G^\star)
-\sum_k\pi_k\mathcal R_k(\boldsymbol\theta_k^\star)\\
&=\sum_k\pi_k(\boldsymbol\theta_G^\star-\boldsymbol\theta_k^\star)^\top
H_k(\boldsymbol\theta_G^\star-\boldsymbol\theta_k^\star).
\end{aligned}
\end{equation}
This statement concerns a population least-squares problem on the feasible affine subspace. Singular cases can be treated by restricting to identifiable directions or using pseudoinverses. The empirical gate adds ridge, shrinkage, clipping, and calibration. For the unconstrained affine population problem, the stated identity is not an out-of-sample performance guarantee for those additions.

Raw ridge-fitted Brier scores are 0.075101842 for global and 0.074064247 for regime affine pooling, a difference of 0.001037595. The reported nine-regime raw-routing diagnostic has correlation 0.960596, slope 0.992201, and mean absolute error 0.000294. These statistics describe theory-aligned pattern agreement and are not an independent proof of the population identity. A forecast-round-blocked event-level regime permutation test uses 1,000 permutations. Six null values equal or exceed the observed statistic, giving $(1+6)/1001=0.006993$.

\section{Regime-Level Confidence Intervals}
\label{app:regimes}
\begin{figure}[t]
\centering
\includegraphics[width=\columnwidth]{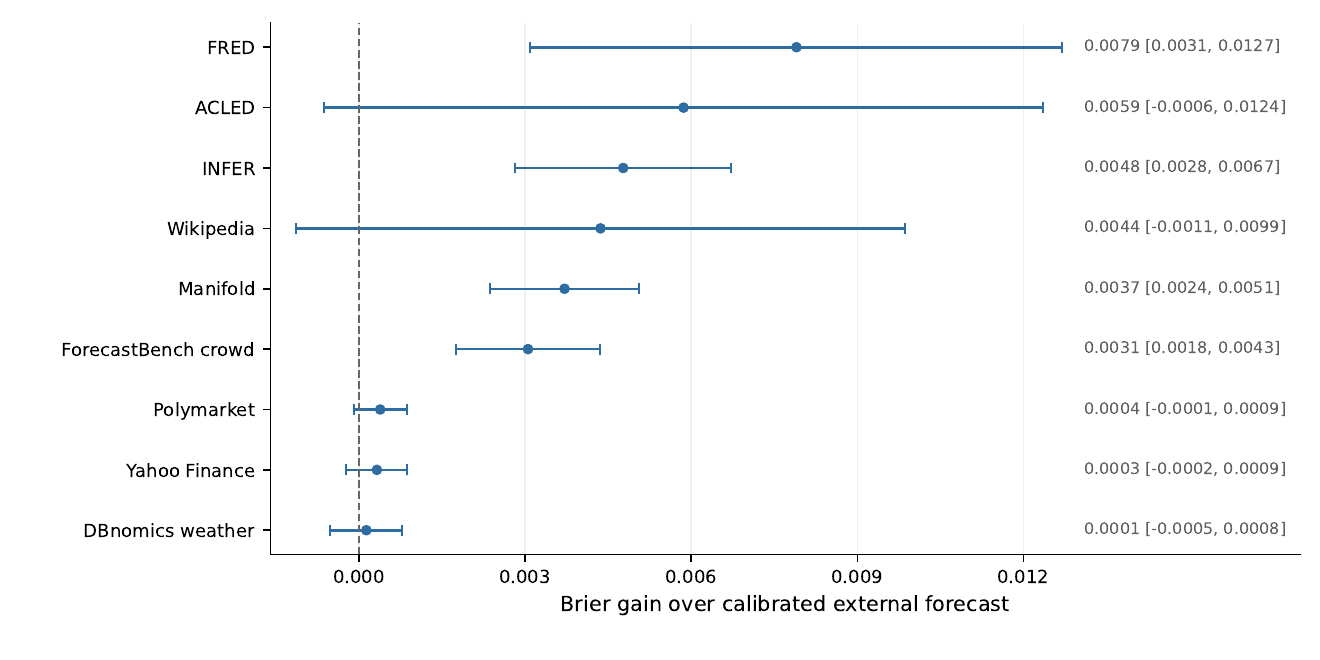}
\caption{Hybrid Brier gain over the calibrated external forecast across nine source-defined regimes. Points show paired mean improvements and horizontal bars show reported 95\% event-clustered confidence intervals. The vertical dashed line indicates no improvement. The ForecastBench crowd label identifies the dataset distribution source while retaining its upstream crowd regime.}
\label{fig:app_source_improvement}
\end{figure}

Figure~\ref{fig:app_source_improvement} and Table~\ref{tab:regime_ci} report source-level gains with event-clustered 95\% confidence intervals. Intervals exclude zero for Manifold, the ForecastBench-derived crowd subset, INFER, and FRED. Several smaller-source intervals include zero.

\begin{table}[t]
\centering\small
\setlength{\tabcolsep}{3.2pt}
\begin{tabular}{lrrr}
\hline
\rowcolor{blue!10}
Regime & Gain & 95\% low & 95\% high\\
\hline
Polymarket & 0.00038 & -0.00010 & 0.00086\\
Manifold & 0.00371 & 0.00236 & 0.00506\\
ForecastBench crowd & 0.00305 & 0.00175 & 0.00435\\
INFER & 0.00477 & 0.00282 & 0.00672\\
FRED & 0.00790 & 0.00309 & 0.01269\\
ACLED & 0.00586 & -0.00064 & 0.01236\\
Wikipedia & 0.00436 & -0.00114 & 0.00986\\
DBnomics & 0.00013 & -0.00052 & 0.00078\\
Yahoo Finance & 0.00032 & -0.00023 & 0.00087\\
\hline
\end{tabular}
\caption{Absolute paired Brier gains against the calibrated external reference, with reported event-clustered 95\% confidence intervals. Several small-source intervals contain zero.}
\label{tab:regime_ci}
\end{table}

\section{Participation and Model Contribution}
\label{app:participation}

For each question, $\lambda_i^{\mathrm{eff}}=\sum_{m=1}^M\tilde w_{k_i,i,m}=1-\tilde w_{k_i,i,g}$ is the \emph{net signed} model-side coefficient. It is not the sum of absolute weight magnitudes and need not lie in $[0,1]$. The bin estimate is the arithmetic mean of the chronological coefficients used for questions in that bin. The later-held-out scalar diagnostic in Table~\ref{tab:participation} has weighted mean gain 0.00104. For each participation quartile $b$, we split questions chronologically within each source, using the earliest 80\% for fitting and latest 20\% for evaluation, then pool the fit rows across sources in that quartile. With $\bar f_i=\frac14\sum_{m=1}^4 f_{im}$, $d_i=\bar f_i-g_i$, and $e_i=g_i-y_i$, we fit $\hat\lambda_b=-\sum_{i\in\mathrm{fit},b}d_ie_i/\sum_{i\in\mathrm{fit},b}d_i^2$. On the later-held-out rows, we predict $\hat h_i=g_i+\hat\lambda_b d_i$ and report $\mathrm{Gain}_b=(N_b^{\mathrm{test}})^{-1}\sum_{i\in\mathrm{test},b}[(g_i-y_i)^2-(\hat h_i-y_i)^2]$. The scalar fit uses earlier forecast dates and is separate from the deployed multi-model gate. Its weighted mean gain of 0.00104 is not directly comparable to the full gate gain of 0.00229 on market and crowd questions. Market and crowd participation proxies are converted to within-source percentile ranks using only snapshot-time information. The 4,454-question market and crowd set yields four pooled equal-rank bins with counts 1,114, 1,114, 1,113, and 1,113.

\begin{table}[t]
\centering\small
\setlength{\tabcolsep}{3.3pt}
\begin{tabular}{lrrr}
\hline
\rowcolor{blue!10}
Quartile & $n$ & $\hat\lambda^{\mathrm{eff}}$ & Gain\\
\hline
Q1, lowest & 1,114 & 0.201794 & 0.001986\\
Q2 & 1,114 & 0.148566 & 0.001330\\
Q3 & 1,113 & 0.083382 & 0.000700\\
Q4, highest & 1,113 & 0.028601 & 0.000146\\
\hline
\end{tabular}
\caption{Mean chronological net model-side coefficients and gains from a separate later-held-out scalar diagnostic by market or crowd participation rank. Gains are evaluated on the later 20\% of each bin against the raw external forecast. The $n$ column reports full-bin counts. These scalar gains are not the deployed multi-model gate gains. Participation is not a direct measure of external accuracy.}
\label{tab:participation}
\end{table}
The Q1 minus Q4 coefficient difference is 0.173193 with approximate 95\% interval [0.09820, 0.24819]. Gain intervals from Q1 through Q4 are [0.001201, 0.002771], [0.000612, 0.002048], [0.000081, 0.001319], and [-0.000294, 0.000586]. A source-fixed-effects regression additionally controls for horizon and probability extremity. Its reported negative association has approximately $p=0.001$. Since participants are not randomized and accuracy does not define the bins, no causal claim follows.

\section{Comparative Confidence and Selectivity}
\label{app:confidence}

\boldheader{Historical Comparative Signals Outperform Raw Confidence}
We evaluate whether confidence and uncertainty signals identify when an individual LM outperforms the raw external forecast. For model $m$, define
\begin{equation}
\label{eq:question_advantage}
z_{im}
=
\mathbf1\{(f_{im}-y_i)^2<(g_i-y_i)^2\}.
\end{equation}
A ridge-logistic estimator with regime fixed effects predicts comparative success from previously resolved examples. Its penalty is fixed at 0.01, and signal orientation is learned from history rather than manually reversing below-chance raw AUCs.

Raw verbal confidence produces AUCs between 0.463 and 0.501 across the primary models. Disagreement and explicit comparative self-assessment obtain raw AUCs of 0.589 and 0.594. Historical comparative competence, evaluated over model--question pairs using the fitted probabilities $\widehat{\Pr}(z_{im}=1\mid\phi_{im},k_i,\mathcal H_i)$, achieves an AUC of 0.681. For question-level selective ranking, we average these model-specific probabilities:
\begin{equation}
\label{eq:question_competence}
c_i=\frac{1}{M}\sum_{m=1}^{M}
\widehat{\Pr}(z_{im}=1\mid
\phi_{im},k_i,\mathcal H_i),
\end{equation}
where $\phi_{im}$ contains the comparative signals available for model $m$ at forecast time. The resulting score measures historical comparative success, not expected hybrid gain. Table~\ref{tab:confidence_auc} reports the model--question-pair AUC comparisons.

\begin{table}[t]
\centering\small
\setlength{\tabcolsep}{4pt}
\begin{tabular}{lrr}
\hline
\rowcolor{blue!10}
Signal & Raw AUC & Historical AUC\\
\hline
20b verbal confidence & 0.488 & 0.548\\
120b verbal confidence & 0.463 & 0.572\\
Llama 8b verbal confidence & 0.501 & 0.533\\
Llama 70b verbal confidence & 0.474 & 0.560\\
Entropy or sharpness & 0.552 & 0.562\\
Model disagreement & 0.589 & 0.608\\
Comparative self-assessment & 0.594 & 0.611\\
Historical competence (pairs) & -- & 0.681\\
\hline
\end{tabular}
\caption{Raw and past-only historically fitted AUC for predicting standalone model wins over the raw external forecast, evaluated over model--question pairs. The historical fit includes regime effects. The averaged question-level score is used separately for selective ranking.}
\label{tab:confidence_auc}
\end{table}

We separately test whether historical competence identifies realized hybrid improvement. Define
\begin{equation}
\label{eq:relative_hybrid_gain}
r_i=
(g_i^{\mathrm{cal}}-y_i)^2-(h_i-y_i)^2.
\end{equation}
At each coverage level, we average the signed relative loss $\ell_i^{\mathrm{rel}}=-r_i$ over retained questions and integrate the resulting risk-coverage curve. Lower relative AURC indicates greater hybrid improvement.

The relative AURCs are $-0.002538$ (random), $-0.002946$ (history-oriented verbal), $-0.003718$ (comparative self-assessment), $-0.005024$ (question-level competence), and $-0.006358$ (oracle). The competence ranking improves on verbal ranking by 0.002078 AURC (95\% CI, [0.001381, 0.002775]).

Individual-model success against the raw external forecast does not guarantee an optimal ranking of hybrid gains over the calibrated external forecast. In particular, ranking by $\Pr(z_{im}=1)$ is not generally optimal for ranking $\mathbb E[r_i\mid\text{features}]$. Comparative self-assessment is a separate diagnostic and does not feed the gate. Unlike the primary question-only forecasts, the diagnostic explicitly provides the model with the external probability $g_i$. The prompt is ``Your forecast is $f_{im}$, while an external source forecasts $g_i$. Before the outcome is known, estimate the probability that your forecast will have lower squared error than the external forecast.'' We therefore do not interpret self-assessment as independent of the external forecast.

\section{Statistical Inference and Locked Holdout}
\label{app:robustness}
Let $d_i=(g_i^{\mathrm{cal}}-y_i)^2-(h_i-y_i)^2$. The primary mean paired gain is 0.002538342, with reported event-cluster standard error 0.000598700, $z=4.2398$, two-sided $p=2.24\times10^{-5}$, and 95\% interval [0.001364890, 0.003711794]. The fixed-comparator Brier skill is 3.3520\%. We resample event IDs 10,000 times with replacement and carry all constituent question snapshots in each draw. The bootstrap distribution supplies the event-cluster standard error. The reported 95\% interval is the Wald interval $\hat\Delta_1\pm1.96\widehat{\mathrm{SE}}_{\mathrm{cluster}}$, and the two-sided $p$-value uses $z=\hat\Delta_1/\widehat{\mathrm{SE}}_{\mathrm{cluster}}$ under the normal approximation. These are not percentile-bootstrap intervals.

The calibrated routing difference is $0.074279618-0.073187332=0.001092286$, with reported standard error 0.000389500, 95\% interval [0.000328866, 0.001855706], and $p=0.00504$. The routing result is secondary because its 95\% interval includes gains below the prespecified 0.001 Brier smallest interesting effect. The estimated minimum detectable effect is 0.0010906.

Development ends on December 31, 2025. The protocol excludes the following eight weeks, and locked testing begins on February 26, 2026. The event-disjoint locked test has 1,173 questions in 947 clusters. Its gain is 0.002269 with 95\% interval [-0.000401, 0.004939]. Using development dispersion $\sigma_d\approx0.0384242$ and the reported effect, the approximate required size is $(2.8\sigma_d/\Delta_1)^2\approx1,796.5$ clusters. The available holdout has insufficient clusters for that target. The confidence interval crosses zero, so this result is not independent confirmation of improvement.

\section{Reproducibility and Artifacts}
\label{app:provenance}
The execution environment uses Linux on x86-64, Python 3.11, vLLM 0.25.1 for LLM inference, PyTorch 2.11.0, CUDA 12.9, and bfloat16 inference on NVIDIA A100-SXM4 GPUs with 80 GB memory per GPU. The host uses an AMD EPYC 7763 CPU and 2 TB system RAM. Forecast aggregation and evaluation use scikit-learn 1.7.2 and XGBoost 3.0.5. The stochastic baseline fitting and resampling configuration uses seed 42 where applicable. Primary LLM predictions use temperature zero. Statistical uncertainty uses 10,000 event-cluster bootstrap samples and the shuffled-regime analysis uses 1,000 blocked event-level permutations.

The reproduction package records population manifests, point-in-time source snapshots, ALFRED vintage metadata, question and prompt hashes, model checkpoints, parsed probabilities, historical out-of-sample calibration inputs, exact baseline configurations, and clustered inference scripts. The structured forecast records identify component probabilities, the number of available components, and use of the historical fallback. The contextual comparator settings, six-dimensional local pooling variables, historical scaling, caliper grid, local log-score weighting, and logistic-stacking feature transformations are specified in Appendix~\ref{app:baselines}.



\end{document}